\documentclass[10pt,twocolumn,letterpaper]{article}
\usepackage{cvpr}
\usepackage{times}
\usepackage{subfigure}
\usepackage{graphicx}
\usepackage{amsmath}
\usepackage{amssymb}
\usepackage{amsthm}
\usepackage[linesnumbered,boxed]{algorithm2e}
\renewcommand{\vec}[1]{\mbox{\boldmath ${#1}$}}
\newcommand{\Matrix}[1]{\mbox{\boldmath ${#1}$}}

\cvprfinalcopy
\def\cvprPaperID{709} % *** Enter the CVPR Paper ID heregoo
\begin{document}
\def\httilde{\mbox{\tt\raisebox{-.5ex}{\symbol{126}}}}
%%%%%%%%% TITLE
\title{Learning Fine-grained Image Similarity with Deep Ranking}
\author{Jiang Wang\textsuperscript{1}\thanks{The work was performed while Jiang Wang and Bo Chen interned at Google.} \quad Yang Song\textsuperscript{2} \quad Thomas Leung\textsuperscript{2}  \quad  Chuck Rosenberg\textsuperscript{2} \quad  Jingbin Wang\textsuperscript{2} \\
James Philbin\textsuperscript{2} \quad Bo Chen\textsuperscript{3} \quad Ying Wu\textsuperscript{1}\\
\textsuperscript{1}Northwestern University \quad \textsuperscript{2}Google Inc. \quad \textsuperscript{3}California Institute of Technology \\
{\footnotesize \texttt{jwa368,yingwu@eecs.northwestern.edu} \quad \texttt{yangsong,leungt,chuck,jingbinw,jphilbin@google.com} \quad
\texttt{bchen3@caltech.edu}}
}
\maketitle
 \thispagestyle{empty}
 \begin{abstract}
 Learning  fine-grained image similarity is a challenging task.
It needs to capture between-class and within-class image differences. This paper proposes a deep ranking model that employs deep learning techniques
  to learn similarity metric directly from images.
It has higher learning capability than models based on hand-crafted features.
A novel multiscale network structure has been developed to describe the images effectively.
An efficient triplet sampling algorithm is proposed to learn the model with distributed asynchronized stochastic gradient. Extensive experiments
  show that the proposed algorithm outperforms models based on
  hand-crafted visual features and deep classification models.
 \end{abstract}

\section{Introduction}
Search-by-example, i.e. finding images that are similar to
a query image, is  an indispensable function for modern image search engines. An effective
image similarity metric is at the core of finding similar images.

Most existing image similarity models consider {\em category-level} image similarity.
For example, in \cite{hadsell2006dimensionality, taylor2011learning}, two images are considered similar as long as they
belong to the same category.
This  category-level image similarity is not sufficient for the search-by-example image
search application.  Search-by-example requires the distinction of differences between images within the same category,
i.e., {\em fine-grained image similarity}.

One way to build image similarity models is to first  extract features
like Gabor filters, SIFT~\cite{lowe1999object} and HOG~\cite{Dalal}, and then learn the
image similarity models on top of these features ~\cite{boureau2010learning, chechik2010large,  taylor2011learning}.
The performance of these methods is  largely limited
by the representation power of the hand-crafted features. Our extensive evaluation has verified that
 being able to jointly learn the features and similarity models with supervised similarity information provides great potential for more effective fine-grained image similarity models than hand-crafted features.

\begin{figure}
  \begin{center}
    \includegraphics[width=8cm]{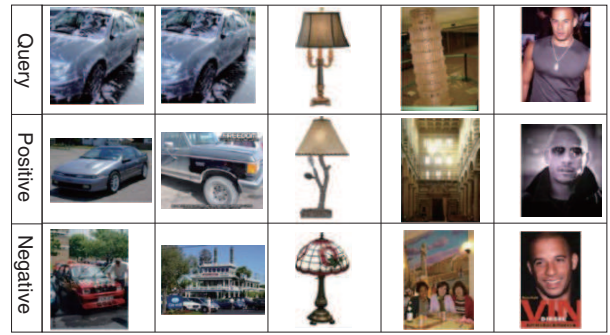}
  \end{center}
  \caption{Sample images from the triplet dataset. Each column is
a  triplet. The upper, middle and lower rows correspond to
query image, positive image, and negative image, where the positive
image is more similar to the query image that the negative image, according to the human raters.
The data are available at {https://sites.google.com/site/imagesimilaritydata/}.}
  \label{fig:sample_triplets}
\end{figure}

Deep learning models
have achieved great success on image classification tasks~\cite{krizhevsky2012imagenet}.
However, similar image ranking is different from image classification.  For image classification, ``black car'', ``white car'' and ``dark-gray car'' are all cars,
while for similar image ranking, if a query image is a ``black car'', we usually want to rank the ``dark gray car'' higher than the ``white car''.
We postulate that image classification models may not fit directly to task of  distinguishing fine-grained image similarity. This hypothesis is verified in experiments. In this paper, we propose  to learn fine-grained image similarity with a {\em deep ranking} model,
which characterizes the fine-grained image similarity relationship with a set of triplets.
A triplet contains a query image, a positive image, and a negative image, where the positive image is  more similar to the query image than the negative image (see Fig. \ref{fig:sample_triplets} for an illustration).
The image similarity relationship is characterized by relative similarity ordering in the triplets.
Deep ranking models can employ this fine-grained image similarity information, which is not considered in category-level image similarity models or classification models, to achieve better performance.

As with most machine learning problems, training data is critical for learning fine-grained image similarity. It is challenging to collect large data sets, which is required for training deep networks. We propose a novel “bootstrapping” method (section \ref{sec:training}) to generate training data, which can virtually generate unlimited amount of training data. To use the data efficiently,  an online triplet sampling algorithm is proposed to generate meaningful and discriminative {\em triplets}, and to utilize asynchronized stochastic gradient algorithm in optimizing triplet-based ranking function.

The impact of different network structures on similar image ranking is explored. Due to the intrinsic difference between image classification and similar image ranking tasks,  a good network for image classification (~\cite{krizhevsky2012imagenet}) may not be optimal for distinguishing fine-grained image similarity. A novel multiscale network structure has been developed, which contains the convolutional neural network with two low resolution paths.  It is shown that this multi-scale network structure can work effectively for similar image ranking.

The image similarity models are evaluated on a human-labeled dataset.
Since it is error-prone for human labelers to directly label the image ranking which may consist tens of images,
we label the similarity relationship of the images with triplets, illustrated in Fig. \ref{fig:sample_triplets}.
The performance of an image similarity model is determined by the fraction of the triplet orderings that agrees with the
ranking of the model. To our knowledge, it is the first high quality dataset with similarity ranking information for images from the same category. We compare the proposed deep ranking model with state-of-the-art methods on this dataset.
The experiments show that the deep ranking model outperforms the hand-crafted visual feature-based approaches~\cite{lowe1999object,Dalal,chechik2010large, taylor2011learning}
and deep classification models~\cite{krizhevsky2012imagenet} by a large margin.

The main contributions of this paper includes the following.
(1) A novel deep ranking model that can learn fine-grained image similarity model directly from images is proposed. We also propose a new “bootstrapping” way to generate the training data. (2) A multi-scale network structure has been developed. (3) A computationally efficient online triplet sampling algorithm is proposed, which is essential for learning deep ranking models with online learning algorithms. (4) We are publishing an evaluation dataset. To our knowledge, it is the first public data set with similarity ranking information for images from the same category (Fig. \ref{fig:sample_triplets}).

\section{Related Work}
Most prior work on image similarity learning~\cite{wang2009learning, guillaumin2009tagprop} studies the category-level image similarity,
where two images are considered similar as long as they belong to the same category.
Existing deep learning models for image similarity also focus on learning category-level image similarity~\cite{taylor2011learning}.
Category-level image similarity mainly corresponds to semantic similarity. \cite{deselaers2011visual} studies the relationship between visual similarity and semantic similarity. It shows that although visual and semantic similarities are generally consistent with each other across different categories,  there still exists considerable   visual variability within a category, especially when the category's semantic scope is large. Thus, it is worthwhile to
learn a fine-grained model that is capable of characterizing the fine-grained visual similarity for the images within the same category.

The following works are close to our work in the spirit of  learning fine-grained image similarity. Relative attribute~\cite{parikh2011relative} learns image attribute ranking among  the images with the same attributes. OASIS~\cite{chechik2010large} and local distance learning~\cite{frome2006image}
learn fine-grained image similarity ranking models on top of the hand-crafted features. These above works are not deep learning based.
\cite{wu2013online} employs deep learning architecture to learn ranking model, but it learns deep network from the ``hand-crafted features'' rather than directly from the pixels.
 In this paper, we propose a Deep Ranking model, which integrates the deep learning techniques
and fine-grained ranking model to learn fine-grained image similarity ranking model directly from images.
The Deep Ranking models perform much better than category-level image similarity models in image retrieval applications.

Pairwise ranking model is a widely used learning-to-rank formulation. It is used to learn image ranking models in \cite{chechik2010large, parikh2011relative,frome2006image}. Generating good triplet samples is a crucial aspect of learning pairwise ranking model.
In \cite{chechik2010large} and  \cite{parikh2011relative}, the triplet sampling algorithms assume that we can load the whole dataset into memory, which is impractical for a large dataset. We design a computationally efficient online triplet sampling
algorithm that does not require loading the whole dataset into memory, which
makes it possible to learn deep ranking models with very large amount of training data.

\section{Overview}
Our goal is to learn image similarity models.
We define the similarity of two images $P$ and $Q$ according
to  their squared Euclidean distance in the image embedding space:
\begin{equation}\label{eq:distance}
D(f(P), f(Q)) = \|f(P) - f(Q) \|_2^2
\end{equation}
where $f(.)$ is the image embedding function that maps an image to a point in an Euclidean space, and
$D(., .)$ is  the squared Euclidean distance in this space. The smaller the distance $D(P, Q)$ is, the more similar the two images $P$ and $Q$ are.
This definition formulates the similar image ranking problem as nearest neighbor search problem in Euclidean space,
which can be efficiently solved via approximate nearest neighbor search algorithms.

We employ the pairwise ranking model to learn image similarity ranking models, partially motivated by \cite{chechik2010large, parikh2011relative}.
Suppose we have a set of images $\mathcal P$, and $r_{i,j} = r(p_i, p_j)$ is a pairwise relevance score which states how similar
the image $p_i \in \mathcal P$ and $p_j \in \mathcal P$ are. The more similar two images are, the higher their relevance
score is.
Our goal is to learn an embedding function $f(.)$ that assigns smaller distance to more similar image pairs, which can be expressed as:
\begin{equation}\label{eq:criteria}
  \begin{aligned}
  &D(f(p_i), f(p_i^+)) <   D(f(p_i), f(p_i^-)), \\
  &\forall \text{$p_i, p_i^+, p_i^-$ such that  $r(p_i, p_i^+) > r(p_i, p_i^-)$}
  \end{aligned}
\end{equation}
We call $t_i = (p_i, p_i^+, p_i^-)$ a triplet, where $p_i, p_i^+, p_i^-$ are the query image, positive image, and negative image, respectively.
A triplet characterizes a relative similarity ranking order for the images $p_i, p_i^+, p_i^-$.
We can define the following hinge loss for a triplet: $t_i = (p_i, p_i^+, p_i^-)$:
\begin{equation}\label{eq:hinge_loss}
  \begin{aligned}
  &l(p_i, p_i^+, p_i^-) = \\
   &\max \{0, g + D(f(p_i), f(p_i^+)) - D(f(p_i), f(p_i^-)) \}
  \end{aligned}
\end{equation}
where $g$ is a gap parameter that regularizes the gap between the distance of the two
image pairs: $(p_i, p_i^+)$ and  $(p_i, p_i^-)$. The hinge loss is a convex
approximation to the 0-1 ranking error loss, which measures the model's violation of the ranking order specified in
the triplet.
Our objective function is:
\begin{equation}\label{eq:objective_loss}
  \begin{aligned}
    \min  &\sum_i \xi_i  + \lambda \| \Matrix W \|_2^2     \\
    s.t.: & \max \{0, g + D(f(p_i), f(p_i^+)) - D(f(p_i), f(p_i^-)) \} \le \xi_i \\
       & \forall \text{$p_i, p_i^+, p_i^-$ such that  $r(p_i, p_i^+) > r(p_i, p_i^-)$}
     \end{aligned}
\end{equation}
where $\lambda$ is a regularization parameter that controls the margin of the learned ranker to improve
its generalization. $\Matrix W$ is the parameters of the embedding function $f(.)$. We employ $\lambda = 0.001$ in this paper. \eqref{eq:objective_loss}
can be converted to an unconstrained optimization by replacing $\xi_i = \max \{0, g + D(f(p_i), f(p_i^+)) - D(f(p_i), f(p_i^-)) \}$.

In this model, the most crucial component is to learn an image embedding function $f(.)$. Traditional
methods typically employ hand-crafted visual features, and learn linear or
nonlinear transformations to obtain the image embedding function. In this paper, we employ the deep learning technique
to learn image similarity models directly from images. We will describe the network architecture of the triple-based ranking loss function in \eqref{eq:objective_loss} and an efficient optimization algorithm to minimize this objective function in the following sections.

\section{Network Architecture}
\label{sec:network}
\begin{figure}
  \begin{center}
    \includegraphics[width=4cm]{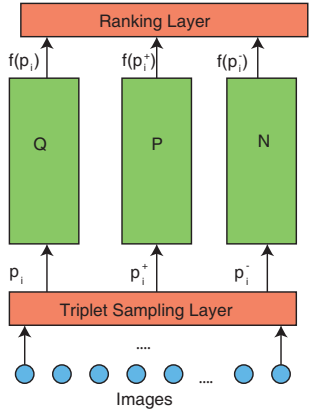}
  \end{center}
  \caption{The network architecture of deep ranking model.}
  \label{fig:network_structure}
\end{figure}

A triplet-based network architecture is proposed for the ranking loss function \eqref{eq:objective_loss}, illustrated
in Fig. \ref{fig:network_structure}.
This network takes image triplets as input. One image triplet contains a query image $p_i$, a positive image
$p_i^+$ and a negative image $p_i^-$, which are fed independently into three identical deep neural networks $f(.)$ with shared architecture and
parameters.  A triplet characterizes the relative similarity relationship for the three images.
The deep neural network $f(.)$ computes the embedding of an image $p_i$: $f(p_i) \in \mathcal R^d$,
where $d$ is the dimension of the feature embedding.

A {\em ranking layer} on the top evaluates the hinge loss \eqref{eq:hinge_loss} of a triplet. The ranking layer does not have any parameter. During
learning, it evaluates the model's violation of the ranking order, and back-propagates the gradients to the lower layers so that
the lower layers can adjust their parameters to minimize the ranking loss \eqref{eq:hinge_loss}.

\begin{figure}
  \begin{center}
    \includegraphics[width=9cm]{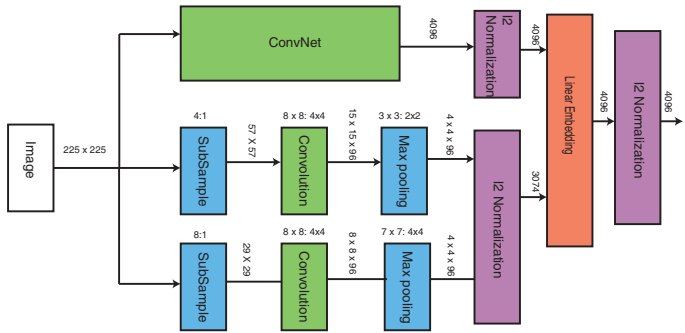} % use a better
                                % drawing.
    % \includegraphics{1}
  \end{center}
  \caption{The multiscale network structure.
    Ech input image goes through three paths.
    The top green box (ConvNet) has the same architecture as
    the deep convolutional neural network in
    \cite{krizhevsky2012imagenet}. The bottom parts are two
    low-resolution paths that extracts low resolution visual
    features. Finally, we normalize the features from both parts, and
    use a linear embedding to combine them. The number shown on the top of a
    arrow is the size of the output image or feature. The number shown on the top of
    a box is the size of the kernels for the corresponding layer.}
  \label{fig:multilevel_network}
\end{figure}

We design a novel multiscale deep neural network architecture  that employs different levels of
invariance at different scales, inspired by \cite{farabet2013learning}, shown in Fig. \ref{fig:multilevel_network}.
The ConvNet in this figure has the same architecture as the convolutional deep neural network in \cite{krizhevsky2012imagenet}. The ConvNet
encodes strong invariance and captures the image semantics.
The other two parts of the network takes down-sampled images and use shallower network architecture.
Those two parts have less invariance and capture the visual appearance.
Finally, we normalize the embeddings from the three parts, and combine them with a linear embedding layer.
In this paper, The dimension of the embedding is $4096$.

We start with a convolutional network (ConvNet) architecture for each individual network, motivated by the recent success of
ConvNet in terms of scalability and generalizability for image classification~\cite{krizhevsky2012imagenet}.
The ConvNet contains stacked {\em convolutional layers},
{\em max-pooling layer}, {\em local normalization layers} and {\em fully-connected layers}. The readers can refer to
\cite{krizhevsky2012imagenet} or the supplemental materials for more details.

 A {\em convolutional layer} takes an image or the feature maps of another layer as input,
convolves it with a set of $k$ learnable kernels, and puts through the activation function
to generate $k$ feature maps. The convolutional layer can be considered as a set of local feature detectors.

A {\em max pooling layer} performs max pooling over a local neighborhood around a pixel. The max pooling layer makes the feature maps
robust to small translations.

A {\em local normalization layer} normalizes the feature map around a local neighborhood to have unit norm and zero mean.
It leads to feature maps that are robust to the differences in illumination and contrast.

The stacked convolutional layers,
 max-pooling layer and local normalization layers act as translational and contrast robust
local feature detectors. A fully connected layer computes a non-linear transformation from the feature
maps of these local feature detectors.

Although ConvNet achieves very good performance for image classification,
the strong invariance encoded in its architecture can be harmful for fine-grained image similarity tasks.
The experiments show that the multiscale network architecture outperforms single scale ConvNet in fine-grained image similarity task.

\section{Optimization}
Training a deep neural network usually needs a  large amount of
training data, which may not fit into the memory of a single computer.
Thus, we employ the distributed asynchronized stochastic gradient algorithm proposed in~\cite{jeff2012distributed} with
momentum algorithm~\cite{sutskeverimportance}.
The momentum algorithm is a stochastic variant of Nesterov's accelerated gradient method~\cite{nesterov1093accelerated}, which converges faster than
traditional stochastic gradient methods.

Back-propagation scheme is used to compute the gradient. A deep network can be represented as the composition of
the functions of each layer.
\begin{equation}\label{eq:composition}
f(.) = g_n(g_{n-1}(g_{n-2}(\cdots g_1(.) \cdots)))
\end{equation}
where $g_l(.)$ is the forward transfer function of the $l$-th layer.
The parameters of the transfer function $g_l$ is denoted as $\vec w_l$. Then the gradient $\frac{\partial f(.)}{\partial \vec w_l}$
can be written as: $\frac{\partial f(.)}{\partial g_l} \times \frac{\partial g_l}{\partial \vec w_l} $, and  $\frac{\partial f(.)}{\partial g_l}$ can be
efficiently computed in an iterative way: $ \frac{\partial f(.)}{\partial g_{l+1}} \times \frac{\partial g_{l+1}(.)}{\partial g_l}$.
Thus, we only need to compute the gradients $\frac{\partial g_l}{\partial \vec w_l}$ and $\frac{\partial g_l}{\partial g_{l-1}}$
for the function $g_l(.)$.
More details of the optimization can be found in the supplemental materials.

To avoid overfitting, dropout~\cite{hinton2012improving} with keeping probability 0.6 is applied to all the fully connected layers.  Random
pixel shift is applied to the input images for data augmentation.

\subsection{Triplet Sampling}
\label{sec:sampling}
To avoid overfitting, it is desirable to utilize a large variety of images. However,  the number of possible triplets increases cubically with the number of images. It is computationally prohibitive and sub-optimal to use all the triplets.
For example, the training dataset in this paper contains 12 million images. The number of all possible triplets in this dataset
is approximately $(1.2 \times 10^7)^3 = 1.728 \times 10^{21}$. This is an extermely large number that can not be
enumerated. If the proposed triplet sampling algorithm is employed, we find the optimization  converges with about 24 million triplet samples, which is a lot smaller than the number of  possible triplets in our dataset.

It is crucial to choose an effective triplet sampling strategy to select the most important triplets for rank learning.
Uniformly sampling of the triplets is sub-optimal, because we are more interested
in the top-ranked results returned by the ranking model. In this paper, we employ an online importance sampling scheme to sample triplets.

Suppose we have a set of images $\mathcal P$, and their pairwise relevance scores $r_{i,j} = r(p_i, p_j)$.
Each image $p_i$ belongs to a category, denoted by $c_i$.
Let the total relevance score of an image $r_i$ defined as
\begin{equation}\label{eq:total_relevance}
r_i = \sum_{j: c_j = c_i, j \neq i}{r_{i,j}}
\end{equation}
The total relevance score of an image $p_i$ reflects how relevant the image is
in terms of its relevance to the other images in the same category.

To sample a triplet, we first sample a query image $p_i$ from $\mathcal P$ according to its total relevance score.
The probability of an image being chosen as query image is proportional to its total relevance score.

Then, we sample a positive image $p_{i}^+$ from the images sharing the same categories as $p_i$. Since we are more interested in the
top-ranked images, we should sample more positive images $p_{i}^+$ with high relevance scores $r_{i,i^+}$. The probability
of choosing an  image $p_{i}^+$ as positive image is:
\begin{equation}\label{eq:sampling}
 P(p_i^+) = \frac{\min\{T_p, r_{i,i^+}\}}{Z_i}
\end{equation}
where $T_p$ is a threshold parameter, and the normalization constant $Z_i$ equals  $\sum_{i^+} P(p_i^+)$ for all the $p_i^+$ sharing the
the same categories with $p_i$.

We have two types of negative image samples. The first type is {\em out-of-class} negative samples,
which are the negative samples that are in a different category from query image $p_i$.
They are drawn uniformly from all the  images with different categories with $p_i$.
The second type   is {\em in-class} negative samples, which are the
negative samples that are in the same category as $p_i$ but is less relevant to $p_i$ than $p_i^+$.
Since we are more interested in the top-ranked images, we draw in-class negative samples $p_i^-$ with the same distribution as (\ref{eq:sampling}).
In order to ensure robust ordering between $p_i^+$ and $p_i^-$ in a triplet $t_i =(p_i, p_i^+, p_i^-)$,   we
also require that the margin between
the relevance score $r_{i, i^+}$ and $r_{i, i^-}$ should be larger than $T_r$, i.e.,
\begin{equation}\label{eq:margin_relevance}
  r_{i, i^+} - r_{i, i^-} \geq T_r, \forall t_i =(p_i, p_i^+, p_i^-)
\end{equation}
We reject the triplets that do not satisfy this condition. If the number of failure trails for one example exceeds
a given threshold, we simply discard this example.

Learning deep ranking models requires large amount of data, which cannot be loaded into main memory. The  sampling algorithms that require random access to all
the examples in the dataset are not applicable.
In this section, we propose an efficient online triplet sampling algorithm based on reservoir sampling~\cite{efraimidis2010weighted}.

We have a set of buffers to store images. Each buffer has a fixed capacity, and it stores images from the same category.
When we have one new image $p_j$, we compute its key $k_j = u_j ^{(1/r_j)}$,where $r_j$ is its total relevance score defined in \eqref{eq:total_relevance} and
$u_j = \text{uniform}(0, 1)$ is a uniformly sampled number. The buffer corresponding to the image $p_j$'s can be found according to its category $c_j$.
If the buffer is not full, we insert the image $p_j$ into the buffer with key $k_j$. Otherwise, we find the image $p_j'$ with
smallest key $k_j'$ in the buffer. If $k_j > k_j'$, we replace the image $p_j'$ with image $p_j$ in the buffer. Otherwise, the imgage example
$p_j$ is discarded.
If this replacing scheme is employed, uniformly sampling from a buffer is equivalent to drawing samples with probability proportional to the total
relevance score $r_j$.

One image $p_i$ is uniformly sampled from all the images in the buffer of category $c_j$ as the query image.
We then uniformly generate one image $p_i^+$ from all the images in the buffer of category $c_j$, and accept it with probability
$\min(1, r_{i, i+}/r_{i+})$, which corresponds to the sampling probability \eqref{eq:sampling}.
 Sampling is continued until one example is accepted. This image example acts as the positive image.

Finally, we draw a negative image sample.
If we are drawing out-of-class negative image sample, we draw a image $p_i^-$
uniformly from all the images in the other buffers. If we are drawing in-class negative image samples, we use the positive example's
drawing method to generate a negative sample, and accept the negative sample only if it satisfies the margin constraint (\ref{eq:margin_relevance}).
 Whether we sample in-class or out-of-class negative samples is controlled by a out-of-class sample ratio parameter.
An illustration of this sampling method is shown in Fig. \ref{fig:sampling}
The outline of reservoir importance sampling algorithm is shown in the supplemental materials.

\begin{figure}
  \begin{center}
    \includegraphics[width=7cm]{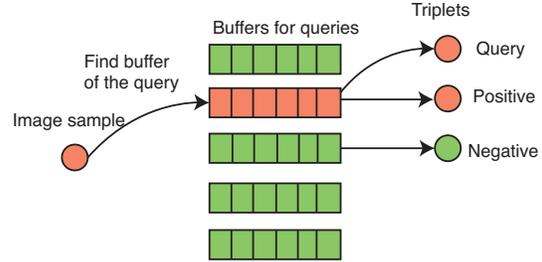}
  \end{center}
  \caption{Illustration of the online triplet sampling algorithm. The negative image in this example is an out-of-class negative. We  have
   one buffer for each category. When we get a new image sample, we insert it into the buffer of the corresponding category
   with prescribed probability. The query and positive examples are sampled from the same buffer, while the negative image is sampled from a different
  buffer.}
  \label{fig:sampling}
\end{figure}
\vspace{-4pt}

\section{Experiments}
\label{sec:experiment}
\subsection{Training Data}
\label{sec:training}
We use two sets of training data to train our model. The first training data is ImageNet ILSVRC-2012 dataset~\cite{imagenet2012},
which contains roughly 1000 images in each of 1000 categories. In total, there are about 1.2 million training images, and 50,000 validation images. This dataset is utilized to learn image semantic information. We use it to pre-train the ``ConvNet'' part of our model using
soft-max cost function as the  top layer.

The second training data is relevance training data, responsible for learning fine-grained visual similarity. The data is generated in a bootstrapping fashion.
It is collected from 100,000 search queries (using Google image search), with the top 140 image results from each query. There are about
14 million images. We employ a {\em{golden feature}}
to compute the relevance $r_{i,j}$ for the images from the same search
query, and set $r_{i,j} = 0$ to the images from different queries.  The golden feature is a weighted linear combination of twenty seven features.
It includes features described in section \ref{sec:visual_features}, with different parameter settings and distance metrics. More importantly, it also includes features learned through image annotation data, such as features or embeddings developed in  \cite{weston2010large}.
The linear weights are learned through max-margin linear weight learning using human rated data.
The golden feature incorporates both visual appearance information and semantic information, and it is of high performance in evaluation. However, it is expensive to compute, and "cumbersome" to develop. This training data is employed to fine-tune our network for fine-grained visual similarity.

\subsection{Triplet Evaluation Data}
Since we are interested in fine-grained similarity, which cannot be characterized by image labels,
we collect a triplet dataset to evaluate image similarity models~\footnote{https://sites.google.com/site/imagesimilaritydata/}.

We started from 1000 popular text queries and sampled triplets $(Q, A, B)$ from the top 50 search results for each query from the Google image search engine. We then rate the images in the triplets using human raters. The raters have four choices: (1) both image A and B are similar to query image Q; (2) both image A and B are dissimilar to query image Q; (3) image A is more similar to Q than B; (4) image B is more similar to Q than A. Each triplet is rated by three raters. Only the triplets with
unanimous scores from the three rates enter the final dataset.
For our application, we discard the triplets with rating (1) and rating (2), because those triplets does not reflect any
image similarity ordering.  About 14,000 triplets are used in evaluation.  Those
triplets are solely used for evaluation. Fig \ref{fig:sample_triplets} shows some triplet examples.

\subsection{Evaluation Metrics}
Two evaluation metrics are used: \emph{similarity precision} and \emph{score-at-top-$K$}  for $K=30$.

Similarity precision is defined as the percentage of triplets being correctly ranked.
Given a triplet $t_i = (p_i, p_i^+, p_i^-)$, where $p_i^+$ should be more similar to $p_i$ than $p_i^-$. Given $p_i$ as query, if $p_i^+$ is ranked
higher than $p_i^-$, then we say the triplet $t_i$ is correctly ranked.

Score-at-top-$K$ is defined as the number of correctly ranked triplets minus the number of incorrectly ranked ones on a subset of triplets
whose ranks are higher than $K$. The subset is chosen as follows.
For each query image in the test set, we retrieve $1000$ images belonging to the same text query, and rank these images using the learned similarity metric.
One triplet's rank is higher than $K$ if its positive image $p_i^+$ or negative image $p_i^-$ is among the top
$K$ nearest neighbors of the query image $p_i$.
This metric is similar to the precision-at-top-$K$ metric, which is widely used to evaluate retrieval systems.
Intuitively, score-at-top-$K$ measures a retrieval system's performance on the $K$ most relevant search results.
This metric can better reflect the performance of the similarity models in practical image retrieval systems,
because users pay most of their attentions to the results on the first few pages.
we set $K=30$ in our experiments.

\subsection{Comparison with Hand-crafted Features}\label{sec:visual_features}
We first compare the proposed deep ranking method with hand-crafted visual features.
For each hand-crafted feature, we report its performance using its best experimental setting. The evaluated hand-crafted visual features include Wavelet~\cite{finkelstein1995fast}, Color (LAB histoghram),
SIFT~\cite{lowe1999object}-like features, SIFT-like Fisher vectors~\cite{perronnin2010large}, HOG~\cite{Dalal},
and SPMK Taxton features with max pooling~\cite{lazebnik2006beyond}. Supervised image similarity ranking information is not used to obtain these features.

Two image similarity models are learned on top of the concatenation of all the visual features described above.\vspace{-6pt}
\begin{itemize}\addtolength{\itemsep}{-0.5\baselineskip}
\item L1HashKCPA~\cite{ioffe2010improved}:  A subset of the golden features (with L1 distance) are chosen using max-margin linear weight learning. We call this set of features ``L1 visual features''.  Weighted Minhash and Kernel principal component analysis (KPCA)~\cite{ioffe2010improved} are applied on the L1 visual features to learn a 1000-dimension embedding in an unsupervised fashion.
\item OASIS~\cite{chechik2010large}: Based on the L1HashKCPA feature, an transformation (OASIS transformation) is learnt with  an online image similarity learning algorithm~\cite{chechik2010large}, using the relevance training data described in Sec. \ref{sec:training}.
\end{itemize}\vspace{-6pt}

The performance comparison is shown in Table \ref{tab:clf}. The ``DeepRanking''  shown in this table is the deep ranking model trained with 20\%
out-of-class negative samples.  We can see that any individual feature without learning does not performs very well.  The L1HashKCPA feature achieves reasonably good performance with relatively low dimension, but its performance is inferior to DeepRanking model.  The  OASIS algorithm can learn better features
because it exploits the image similarity ranking information in the relevance training data.
By directly learning a ranking model on images,
the deep ranking method can use more information from image than two-step
``feature extraction''-``model learning'' approach.
Thus, it performs better both in terms of similarity precision and score-at-top-$30$.

The DeepRanking model performs better in terms of similarity precision than the golden features, which are used to generate relevance training data. This is because
the DeepRanking model employs the category-level information in ImageNet data and relevance training data to better characterize
the image semantics. The score-at-top-$30$ metric of DeepRanking is only slightly lower than the golden features.

\begin{table}[htb]
\begin{center}
    \begin{tabular}{ | c | c | c | }
    \hline
    Method & Precision &   Score-30 \\ \hline
    Wavelet~\cite{finkelstein1995fast} & $62.2\%$ & $2735$ \\
    Color & $62.3\%$  & $2935$ \\
    SIFT-like~\cite{lowe1999object} & $65.5\%$ & $2863$\\
    Fisher~\cite{perronnin2010large}&  $ 67.2\%$ & $3064$\\
    HOG~\cite{Dalal} & $68.4\%$ & $3099$ \\
    SPMKtexton1024max~\cite{lazebnik2006beyond} & $66.5\%$ & $3556$ \\ \hline
    L1HashKPCA~\cite{ioffe2010improved} & $76.2\%$ & $6356$ \\
    OASIS~\cite{chechik2010large} &  $ 79.2\%$ & $6813$\\\hline
    Golden Features & $80.3\%$& ${ \bf 7165}$ \\ \hline
    %CongasFisher (512) &  $ 66.5\%$ & $2983$\\ \hline
    %CongasFisher (2048) &  $64.4\%$ & $2941$\\ \hline
    DeepRanking & ${\bf 85.7}\%$ & $7004$\\ \hline
    \end{tabular}
\end{center}
    \caption{Similarity precision (Precision) and score-at-top-$30$ (Score-30) for different features.}
    \label{tab:clf}
\end{table}
\vspace{-6pt}

\subsection{Comparison of Different Architectures}
We compare the proposed method with the following architectures:
(1) Deep neural network for classification trained on
ImageNet, called ConvNet. This is exactly the same as the model trained in \cite{krizhevsky2012imagenet}.
(2) Single-scale deep neural network
for ranking. It only has a single scale ConvNet in deep ranking model, but  It is trained in the same way as DeepRanking model.
(3) Train an OASIS model~\cite{chechik2010large}  on the feature output of single-scale deep neural network for ranking.
(4) Train a linear embedding on both the single-scale deep neural network and the visual features described in the last section.
The performance are shown in Table \ref{tab:clf_architecture}. In all the experiments, the Euclidean distance of the
embedding vectors of the penultimate layer
before the final softmax or ranking layer is exploited as similarity measure.

First, we find that the ranking model greatly increases the performance. The performance of
single-scale ranking model is much better than ConvNet. The two networks have the same architecture except
single-scale ranking model is fine-tuned with the relevance training data using ranking layer, while ConvNet is trained solely for classification task using logistic regression layer.

We also find that single-scale ranking performs very well in terms of similarity precision,
but its score-at-top-30 is not very high. The DeepRanking model, which employs multiscale
network architecture,  has both better similarity precision and score-at-top-30. Finally, although training
an OASIS model or linear embedding on the top increases performance, their performance is inferior to DeepRanking model,
which uses  back-propagation to fine-tune the whole network.
\begin{table}[htb]
\begin{center}
    \begin{tabular}{ | c | c | c | }
    \hline
    Method & Precision &   Score-30 \\ \hline
    ConvNet & $82.8\%$ & $5772$ \\ \hline
    Single-scale Ranking & $84.6\%$ & $6245$ \\ \hline
    OASIS on Single-scale Ranking & $82.5\%$ & $6263$ \\ \hline
    Single-Scale \& Visual Feature & $84.1 \%$ & $6765$ \\ \hline
    DeepRanking & $ {\bf 85.7}\%$ & ${\bf 7004}$\\ \hline
    \end{tabular}
\end{center}
    \caption{Similarity precision (Precision) and score at top $30$ (Score-30) for different neural network architectures.}
    \label{tab:clf_architecture}
\end{table}
\vspace{-6pt}

\begin{figure}
  \begin{center}
   \includegraphics[width=6cm]{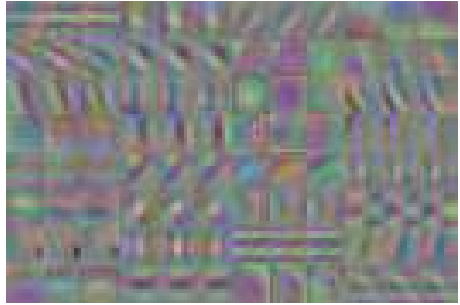}\\
  \end{center}
  \caption{The learned filters of the first level convolutional layers of the multi-scale deep ranking model.}
  \label{fig:filters}
\end{figure}

An illustration of the learned filters of the multi-scale deep ranking model is shown in Fig. \ref{fig:filters}. The filters learned in this paper captures more color information compared with the filter learned in \cite{krizhevsky2012imagenet}.

\vspace{-8pt}

\subsection{Comparison of Different Sampling Methods}
\begin{figure}
  \begin{center}
  \includegraphics[width=4cm]{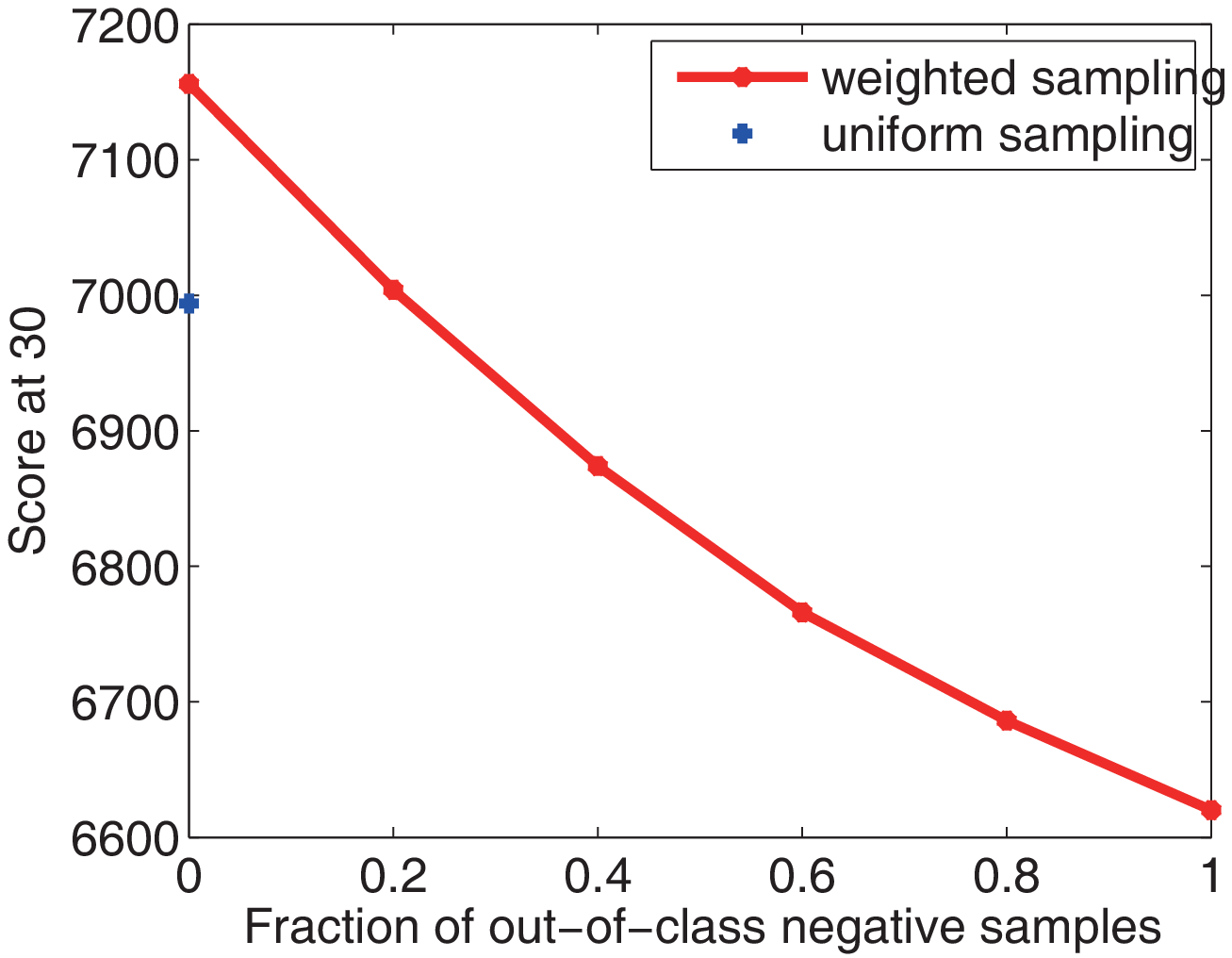}
  \includegraphics[width=4cm]{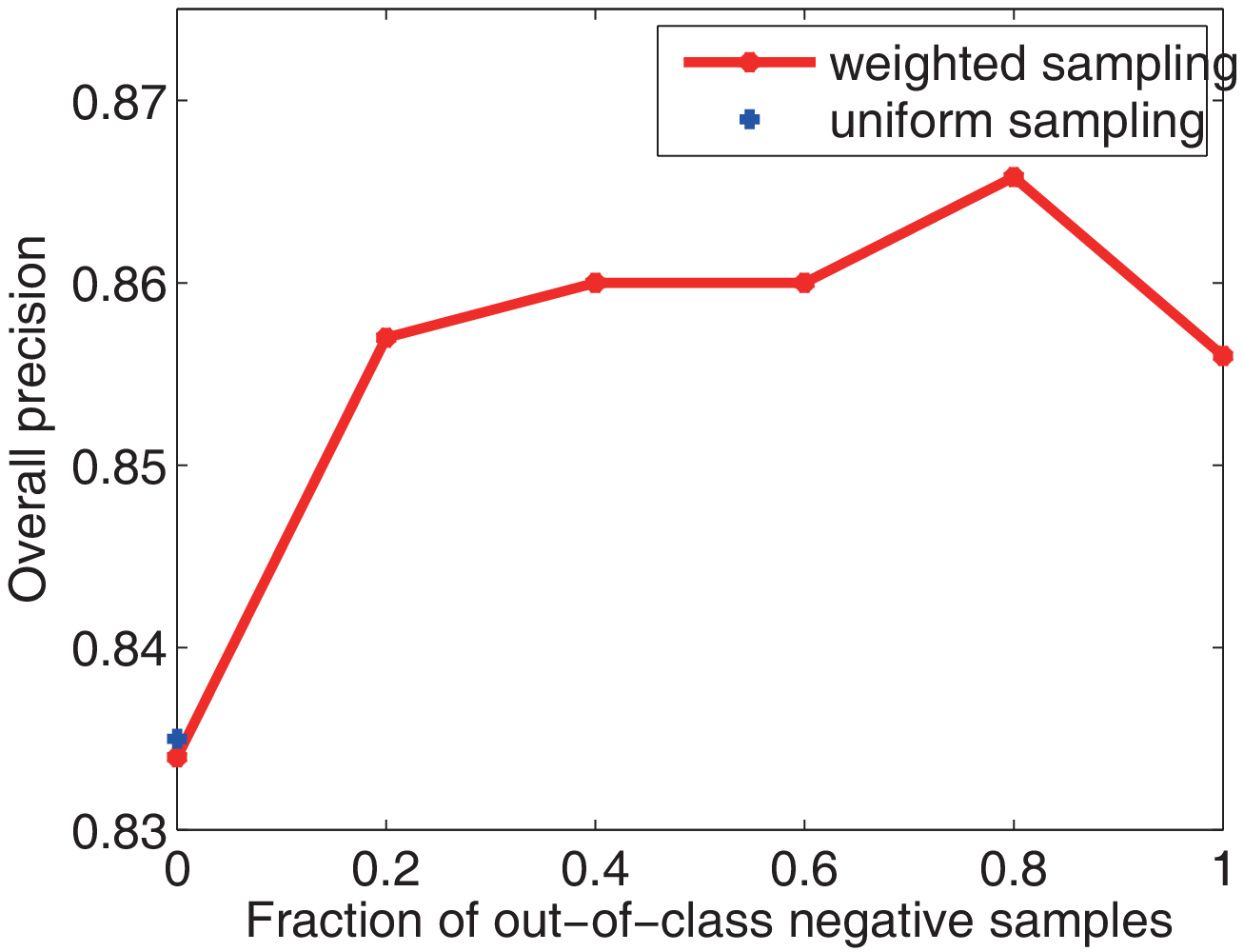}
  \end{center}
  \caption{The relationship between the performance of the proposed method and the fraction of out-of-class negative samples.}
  \label{fig:sample_fraction}
\end{figure}

We study the effect of  the fraction of the out-of-class negative samples in online triplet sampling
algorithm on the performance of the proposed method. Fig. \ref{fig:sample_fraction} shows the results. The results are obtained from drawing 24 million triplets samples.
We find that the score-at-top-30 metric of DeepRanking model decreases as we have more out-of-class negative
samples. However, having a small fraction of out-of-class samples (like 20\%) increases the similarity precision metric a lot.

We also compare the performance of the weighted sampling and uniform sampling with 0\% out-of-class negative samples.
In weighted sampling, the sampling probability of the images is proportional to its total relevance score
$r_j$ and pairwise relevance score $r_{i,j}$, while uniform sampling draws the images uniformly from all
the images (but the ranking order and margin constraints should be satisfied).
We find that although the two sampling methods perform similarly in overall precision, the weighted sampling algorithm
does better in score-at-top-30. Thus, weighted sampling is employed.

\vspace{-6pt}

\subsection{Ranking Examples}
\begin{figure}
  \begin{center}
  \includegraphics[width=8.5cm]{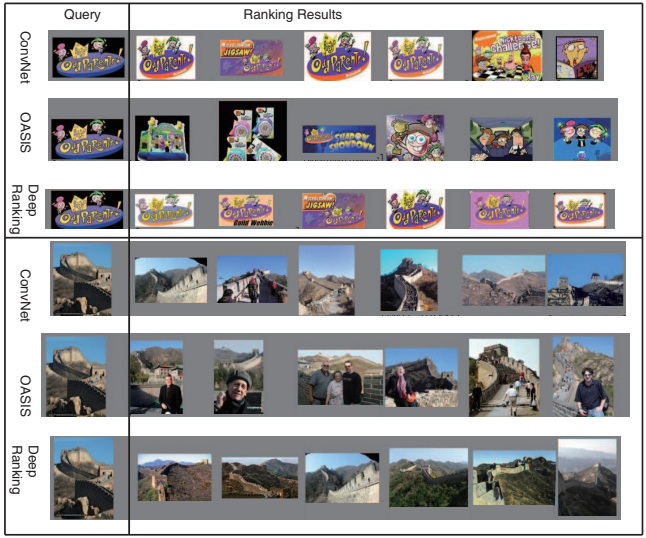}
  \end{center}
  \caption{Comparison of the ranking examples of ConvNet, Oasis Features and Deep Ranking.}
  \label{fig:ranking_examples}
\end{figure}

A comparison of the ranking examples of ConvNet, OASIS feature (L1HashKPCA
features with OASIS learning)  and Deep Ranking is shown in Fig. \ref{fig:ranking_examples}.
We can see that ConvNet captures the semantic meaning of the images very well, but it fails to take into account some global visual appearance, such as
color and contrast. On the other hand, Oasis features can characterize the visual appearance well, but fall short on
the semantics. The proposed deep ranking method incorporates both the visual appearance and image semantics.

\vspace{-10pt}

\section{Conclusion}
In this paper, we propose a novel deep ranking model to learn fine-grained image similarity models.
The deep ranking model employs a triplet-based hinge loss ranking  function to characterize fine-grained
image similarity relationships, and a multiscale neural network architecture to capture both the
global visual properties and the image semantics.
We also propose an efficient online triplet sampling method that enables us to learn  deep ranking
models from very large amount of training data.
The empirical evaluation shows that the deep ranking model achieves much better performance than the state-of-the-art
 hand-crafted features-based models and deep classification models.
Image similarity models can be applied to many other computer vision applications, such as
exemplar-based object recognition/detection and image deduplication. We will explore along these directions.

{\scriptsize
\bibliographystyle{ieee}
\bibliography{../../library_handedited}

\begin{thebibliography}{10}\itemsep=-1pt

\bibitem{imagenet2012}
A.~Berg, D.~Jia, and L.~FeiFei.
\newblock Large scale visual recognition challenge 2012, 2012.

\bibitem{boureau2010learning}
Y.-L. Boureau, F.~Bach, Y.~LeCun, and J.~Ponce.
\newblock Learning mid-level features for recognition.
\newblock In {\em CVPR}, pages 2559--2566. IEEE, 2010.

\bibitem{chechik2010large}
G.~Chechik, V.~Sharma, U.~Shalit, and S.~Bengio.
\newblock Large scale online learning of image similarity through ranking.
\newblock {\em JMLR}, 11:1109--1135, 2010.

\bibitem{Dalal}
N.~Dalal and B.~Triggs.
\newblock {Histograms of Oriented Gradients for Human Detection}.
\newblock In {\em CVPR}, pages 886--893. IEEE, 2005.

\bibitem{jeff2012distributed}
J.~Dean, G.~Corrado, R.~Monga, K.~Chen, M.~Devin, Q.~Le, M.~Mao, M.~Ranzato,
  A.~Senior, P.~Tucker, K.~Yang, and A.~Ng.
\newblock Large scale distributed deep networks.
\newblock In P.~Bartlett, F.~Pereira, C.~Burges, L.~Bottou, and K.~Weinberger,
  editors, {\em Advances in Neural Information Processing Systems 25}, pages
  1232--1240. 2012.

\bibitem{deselaers2011visual}
T.~Deselaers and V.~Ferrari.
\newblock Visual and semantic similarity in imagenet.
\newblock In {\em CVPR}, pages 1777--1784. IEEE, 2011.

\bibitem{efraimidis2010weighted}
P.~S. Efraimidis.
\newblock Weighted random sampling over data streams.
\newblock {\em arXiv preprint arXiv:1012.0256}, 2010.

\bibitem{farabet2013learning}
C.~Farabet, C.~Couprie, L.~Najman, and Y.~LeCun.
\newblock Learning hierarchical features for scene labeling.
\newblock {\em Pattern Analysis and Machine Intelligence, IEEE Transactions
  on}, 35(8):1915--1929, 2013.

\bibitem{finkelstein1995fast}
A.~Finkelstein and D.~Salesin.
\newblock Fast multiresolution image querying.
\newblock In {\em Proceedings of the ACM SIGGRAPH Conference on Visualization:
  Art and Interdisciplinary Programs}, pages 6--11. ACM, 1995.

\bibitem{frome2006image}
A.~Frome, Y.~Singer, and J.~Malik.
\newblock Image retrieval and classification using local distance functions.
\newblock In {\em NIPS}, volume~2, page~4, 2006.

\bibitem{guillaumin2009tagprop}
M.~Guillaumin, T.~Mensink, J.~Verbeek, and C.~Schmid.
\newblock Tagprop: Discriminative metric learning in nearest neighbor models
  for image auto-annotation.
\newblock In {\em ICCV}, pages 309--316. IEEE, 2009.

\bibitem{hadsell2006dimensionality}
R.~Hadsell, S.~Chopra, and Y.~LeCun.
\newblock Dimensionality reduction by learning an invariant mapping.
\newblock In {\em CVPR}, volume~2, pages 1735--1742. IEEE, 2006.

\bibitem{hinton2012improving}
G.~E. Hinton, N.~Srivastava, A.~Krizhevsky, I.~Sutskever, and R.~R.
  Salakhutdinov.
\newblock Improving neural networks by preventing co-adaptation of feature
  detectors.
\newblock {\em arXiv preprint arXiv:1207.0580}, 2012.

\bibitem{ioffe2010improved}
S.~Ioffe.
\newblock Improved consistent sampling, weighted minhash and l1 sketching.
\newblock In {\em Data Mining (ICDM), 2010 IEEE 10th International Conference
  on}, pages 246--255. IEEE, 2010.

\bibitem{krizhevsky2012imagenet}
A.~Krizhevsky, I.~Sutskever, and G.~Hinton.
\newblock Imagenet classification with deep convolutional neural networks.
\newblock In {\em NIPS}, pages 1106--1114, 2012.

\bibitem{lazebnik2006beyond}
S.~Lazebnik, C.~Schmid, and J.~Ponce.
\newblock Beyond bags of features: Spatial pyramid matching for recognizing
  natural scene categories.
\newblock In {\em CVPR}, volume~2, pages 2169--2178. IEEE, 2006.

\bibitem{lowe1999object}
D.~G. Lowe.
\newblock Object recognition from local scale-invariant features.
\newblock In {\em ICCV}, volume~2, pages 1150--1157. IEEE, 1999.

\bibitem{nesterov1093accelerated}
Y.~Nesterov.
\newblock A method of solving a convex programming problem with convergence
  rate o(1/sqr(k)).
\newblock {\em Soviet Mathematics Doklady}, 1983.

\bibitem{parikh2011relative}
D.~Parikh and K.~Grauman.
\newblock Relative attributes.
\newblock In {\em ICCV}, pages 503--510. IEEE, 2011.

\bibitem{perronnin2010large}
F.~Perronnin, Y.~Liu, J.~S{\'a}nchez, and H.~Poirier.
\newblock Large-scale image retrieval with compressed fisher vectors.
\newblock In {\em CVPR}, pages 3384--3391. IEEE, 2010.

\bibitem{sutskeverimportance}
I.~Sutskever, J.~Martens, G.~Dahl, and G.~Hinton.
\newblock On the importance of initialization and momentum in deep learning.
\newblock In {\em ICML}, 2013.

\bibitem{taylor2011learning}
G.~W. Taylor, I.~Spiro, C.~Bregler, and R.~Fergus.
\newblock Learning invariance through imitation.
\newblock In {\em CVPR}, pages 2729--2736. IEEE, 2011.

\bibitem{wang2009learning}
G.~Wang, D.~Hoiem, and D.~Forsyth.
\newblock Learning image similarity from flickr groups using stochastic
  intersection kernel machines.
\newblock In {\em ICCV}, pages 428--435. IEEE, 2009.

\bibitem{weston2010large}
J.~Weston, S.~Bengio, and N.~Usunier.
\newblock Large scale image annotation: learning to rank with joint word-image
  embeddings.
\newblock {\em Machine learning}, 81(1):21--35, 2010.

\bibitem{wu2013online}
P.~Wu, S.~C. Hoi, H.~Xia, P.~Zhao, D.~Wang, and C.~Miao.
\newblock Online multimodal deep similarity learning with application to image
  retrieval.
\newblock In {\em Proceedings of the 21st ACM international conference on
  Multimedia}, pages 153--162. ACM, 2013.

\end{thebibliography}
}

\end{document}